\documentclass{article}

\usepackage{microtype}
\usepackage{subfigure}
\usepackage{booktabs}
\usepackage{amsmath}
\usepackage{amssymb}
\usepackage{mathtools}
\usepackage{amsthm}

\usepackage{bm}
\usepackage{eqnarray}    % provides 'H'
\usepackage{graphicx}    % provides 'subfigure' and corresponding caption

\usepackage{xcolor}
\definecolor{red}{rgb}{1, 0, 0}
\definecolor{blue}{rgb}{0, 0, 1}

\usepackage{enumitem,amssymb}
\newlist{todolist}{itemize}{2}
\setlist[todolist]{label=$\square$}
\usepackage{pifont}
\usepackage[a4paper, total={6.5in, 9in}]{geometry}
\usepackage[colorlinks=true,linkcolor=black,anchorcolor=black,citecolor=blue,filecolor=black,menucolor=black,runcolor=black,urlcolor=black]{hyperref}

\usepackage[square, numbers]{natbib}

\usepackage{tabularx}
\newcolumntype{L}{>{\centering\arraybackslash}m{3cm}}

\usepackage[usestackEOL]{stackengine}

\usepackage{multirow}

\usepackage[width=.875\textwidth]{caption}
\usepackage{adjustbox}

\usepackage{mathpazo}                      % To get Palatino
\usepackage[small, euler-digits]{eulervm}  % and now Euler
\newcommand\raisepunct[1]{\,\mathpunct{\raisebox{0.5ex}{#1}}}

\usepackage[usestackEOL]{stackengine}

\usepackage{authblk}

\begin{document}

\title{Uncertainty in GNN Learning Evaluations: A Comparison Between Measures for Quantifying Randomness in GNN Community Detection.}
\author{William Leeney and Ryan McConville}
\affil{School of Engineering Mathematics and Technology, University of Bristol}

\maketitle

\abstract{(1) The enhanced capability of Graph Neural Networks (GNNs) in unsupervised community detection of clustered nodes is attributed to their capacity to encode both the connectivity and feature information spaces of graphs. The identification of latent communities holds practical significance in various domains, from social networks to genomics. Current real-world performance benchmarks are perplexing due to the multitude of decisions influencing GNN evaluations for this task. (2) Three metrics are compared to assess the consistency of algorithm rankings in the presence of randomness. The consistency and quality of performance between the results under a hyperparameter optimisation with the default hyperparameters is evaluated. (3) The results compare hyperparameter optimisation with default hyperparameters, revealing a significant performance loss when neglecting hyperparameter investigation. A comparison of metrics indicates that ties in ranks can substantially alter the quantification of randomness. (4) Ensuring adherence to the same evaluation criteria may result in notable differences in the reported performance of methods for this task. The $W$ Randomness coefficient, based on the Wasserstein distance, is identified as providing the most robust assessment of randomness.}

\newpage
\section{Introduction}

Graph Neural Networks (GNNs) have gained popularity as a neural network-based approach for handling graph-structured data, leveraging their capacity to merge two information sources through the propagation and aggregation of node feature encodings along the network's connectivity \cite{kipf2016semi}. Nodes within a network can be organized into communities based on similarities in associated features and/or edge density \cite{schaeffer2007graph}. Analyzing the network structure to identify clusters or communities of nodes proves valuable in addressing real-world issues like misinformation detection \cite{monti2019fake}, genomic feature discovery \cite{cabreros2016detecting}, and social network or research recommendation \cite{yang2013community}. We consider unsupervised neural approaches to community detection that do not use any ground truth or labels during training to optimise the loss functions. As an unsupervised task, the identification of node clusters relies on latent patterns within the dataset rather than on ``ground-truth" labels. Clustering holds significance for emerging applications lacking associated ground truth. Evaluating performance in discovering unknown information becomes crucial for applications where label access is restricted. Many graph applications involve millions of nodes, and datasets mimicking realistic scenarios exhibit low labeling rates \cite{hu2021ogb}. 

Recently, \citet{ugle2023leeney} , proposed a framework for fairly evaluating GNN community detection algorithms and the importance of a hyperparameter optimisation procedure to performance comparisons. However, this is not done consistently across the field although benchmarks are widely considered as important. Confusion arising from biased benchmarks and evaluation procedures distorts understanding of the research field. Currently, resources, money, time and energy training models are wasted on inconclusive results which may have developed understanding in the field but go unpublished. Where research findings inform policy decisions or medical practices, publication bias can lead to decisions based on incomplete or biased evidence, in turn creating inefficiencies and downstream harms. To accurately reflect the real-world capabilities of research, one may use a common framework for evaluating proposed methods. To demonstrate the need for such a framework in this setting, we measure the difference between using the default parameters given by the original implementations to those optimised for under this framework. To quantify the influence of randomness on results we compare three metrics, two proposed herein, for evaluating consistency of algorithm rankings over different random seeds which quantifies the robustness of results. Here, we also investigate how the quantification of randomness is affected when ties in performance between two or more algorithms is accounted for. Ties are can occur when out of memory errors occur or when a metric such as conductance is optimised. In addition, we investigate how these metrics change as the number of tests increase, showing how sensitive each is to breathe of investigations.

\subsection{Related Work}

There is a recognition of the necessity for rigor in frameworks assessing machine learning algorithms. \cite{pineau2021improving}.
Several frameworks for assessing the performance of supervised GNNs performing node classification, link prediction and graph classification exist \cite{dwivedi2020benchmarking,morris2020tudataset,errica2019fair,palowitch2022graphworld}. 
Our focus lies in unsupervised community detection, a task that proves more challenging to train and evaluate. While existing reviews on community detection offer insights, they generally lack thorough evaluations. The absence of a standardized evaluation in this task has been acknowledged in discussions concerning non-neural methods \cite{liu2020deep,jin2021survey}, but this consideration does not currently extend to GNNs.

\citet{su2022comprehensive} provides a taxonomy of existing node clustering methods, comparing representative works in the GNN space, as well as deep non-negative matrix factorisation and sparse filtering-based methods but do not empirically evaluate the methods discussed. \citet{chunaev2020community} provides an overview of community detection methods but omit mention of GNN methods. \citet{tu2018unified} provides an overview of network representation learning for community detection and proposes a novel method to incorporate node connectivity and features, but does not compare GNN methods in the study. \citet{ezugwu2022comprehensive} surveys clustering algorithms providing an in-depth discussion of all applications of clustering algorithms, but does not includes an investigation into methods that perform clustering on graphs. There are many traditional community detection methods that are not based on neural networks. Louvain \cite{blondel2008fast} is a classic community detection algorithm that uses two phases: in the modularity optimisation phase, nodes are randomly ordered then added and removed from communities until there is no significant change in modularity; nodes with the same community are collected and represented as a single node in community aggregation. Leiden \cite{traag2019louvain} is an extension of Louvain, designed to fix the tendency to discover weakly connected communities. Leiden has some subtle differences which include the addition of a refinement of partitions phase in-between the modularity optimisation and community aggregation phases of Louvain. Rather than the greedy merging of Louvain based on the largest increase of modularity, the chance of merging increases in proportion to the modularity. Also, Leiden uses a `fast local move procedure', whereby it only visits the nodes who's neighbourhoods have been changed in the last iteration of the modularity optimisation. Label Propagation \cite{raghavan2007near} uses network structure to find communities of densely connected nodes. This works by initialising every node with a unique label and at every iteration, each node is reassigned the label that the majority of it's neighbours have. These traditional methods are transductive \cite{vapnik1998statistical} as they do not learn a function that can be applied to unseen data, whereas in this work we only consider inductive GNN methods which produce a model that can can predict community assignments for data that it has not been trained on. 

Various frameworks exist for evaluating performance, and the evaluation procedure employed significantly influences the performance of all algorithms \cite{zoller2021benchmark}. Under consistent conditions, it has been demonstrated that simple models can exhibit improved performance with a thorough exploration of the hyperparameter space \cite{shchur2018pitfalls}. This improvement may be attributed to the impact of random initialisations on performance \cite{errica2019fair}. Importantly, relying on results from papers without conducting the same hyperparameter optimization across all models introduces inconsistency and yields a misleading benchmark. The biased selection of random seeds, which can skew performance, is considered unfair. Furthermore, not training over the same number of epochs or neglecting model selection based on validation set results leads to unfair comparisons, potentially resulting in inaccurate conclusions about the effectiveness of models. Previous work in this space by \citet{ugle2023leeney} proposed a framework for consistent community detection with GNNs and quantifying the randomness in these investigations. In this work, we expand upon this metric of randomness by considering the affect of ties in performance. We show the affect of ties on ranking randomness and propose two new metrics that improve upon the previous work, evaluating how sensitive each is to the scale of the investigation.

%%%%%%%%%%%%%%%%%%%%%%%%%%%%%%%%%%%%%%%%%%
\section{Methodology}

This section details the procedure for evaluation; the problem that is aimed to solve; the hyperparameter optimisation and the resources allocated to this investigation; the algorithms that are being tested; the metrics of performance and datasets used.

The current method of evaluating algorithms suffer from various shortcomings that impede the fairness and reliability of model comparisons. Establishing a consistent framework provides a transparent and objective basis for comparing models. A standardised benchmark practice contributes to transparency by thoroughly documenting the factors influencing performance, encouraging researchers to engage in fair comparisons. To leverage results from previous research, it is essential to follow the exact evaluation procedure, saving time and effort for practitioners. Establishing consistent practices is crucial, as there is currently no reason for confidence in performance claims without a trustworthy evaluation, fostering a deeper understanding and facilitating progress in the field. For this reason, we follow the procedure established by \citet{ugle2023leeney}.

None of the evaluated algorithms are deterministic, as each relies on randomness for initializing the network. Thus, the consistency of a framework can be assessed by considering the amount to which performance rankings change when different randomness is introduced across all tests within the framework. Tests refer to the metric's performance on a specific dataset, and ranking indicates the algorithm's placement relative to others. In this context, different randomness means each distinct random seed used evaluating the algorithms. To evaluate the consistency of results obtained by this framework, we compare the existing coefficient of randomness with two new metrics. We investigate the affect of performance ties on these metrics. In the existing metrics, ties are dealt with by awarding each algorithm the lowest rank between those that share a rank. In this work, this is compared with the scenario of awarding the mean rank of those that are tied. Ties are likely to occur under certain metrics such as conductance, where the algorithm scores the optimal value of 0. In addition, ties will occur where algorithms run out of memory in computation. With this setup, the improved version of Kendall's $W$ coefficient of concordance\cite{field2005kendall} that can account for ties is used to assess the consistency of rankings. In addition, we use the Wasserstein distance to create another metric for quantifying the difference in rankings due to random seeds. From \citet{ugle2023leeney}, the $W$ randomness coefficient is calculated using the number of algorithms $a$, and random seeds $n$, along with tests of performance that create rankings of algorithms, and defined by Equation~\ref{equ: Worder}:

\begin{equation}
W = 1 - \frac{1}{|\mathcal{T}|}\sum_{t \in \mathcal{T}} \frac{12S}{n^2(a^3-a)} \raisepunct{.}
\label{equ: Worder}
\end{equation}

The sum of squared deviations $S$ is calculated for all algorithms, and calculated using the deviation from mean rank due for each random seed. This is averaged over all metrics and datasets that make up the suite of tests $\mathcal{T}$. Using the one minus means that if the $W$ is high then randomness has affected the rankings, whereas a consistent ranking procedure results in a lower number. By detailing the consistency of a framework across the randomness evaluated, the robustness of the framework can be maintained, allowing researchers to trust results and maintain credibility of their publications. However, this metric does not account for ties in performance and deals with this by assigning the lowest rank from all the algorithms that tie. Instead, an improvement to this is where there are ties, each are given the average of the ranks that would have been given had no ties occurred. Where there are a large number of ties this reduce the value of $W$ and allows us to compute the correction equation as,

\begin{equation}
W_t = 1 -  \frac{1}{|\mathcal{T}|}\sum_{t \in \mathcal{T}}  \frac{12\sum_{i=1}^{a} (R_i^2) - 3n^2a(a+1)^2}{n^2 a(a^2 -1) - n \sum_{j=1}^{n} (\sum_{i=1}^{g_j} (t^{3}_{i} - t_i))}
\raisepunct{,}
\label{equ: Wtieorder}
\end{equation}

where $R_i$ is the sum of the ranks for algorithm $i$, $g_j$ is the number of groups of ties in the rankings under seed $j$ and $t_i$ is the number of ties in that group. 

The other metric that we propose is to calculate the overlap in ranking distributions. This is normalised by the maximum difference that would occur when there is no uncertainty in rank due to randomness as there would be no overlap between the ranking distributions. Formally, we calculate the normalised Wasserstein distance \citep{vallender1974calculation, shen2018wasserstein} between rank distributions over a each of the test scenarios. Therefore, given the probability distribution of the rank of an algorithm $j$ as $f_{j}(r)$ over the discrete ranking space $R$ then the cumulative distribution is denoted as $F_j(r)$ and therefore the Wasserstein distance between two rank distributions is given by Equation~\ref{equ: ww_dist}:

\begin{equation}
 W1(i, j) = \int_{R_a} |F_i(r), F_j(r)| \, dr.
\label{equ: ww_dist}
\end{equation}

This leads us to the definition of the $W_w$ Wasserstein randomness given by Equation~\ref{equ: ww_rand}, 

\begin{equation}
W_w = 1 - \frac{1}{|\mathcal{T}|}\sum_{t \in \mathcal{T}} \frac{\sum_{i=1}^{a} \sum_{j=1}^{i-1} W1(i, j)}{\sum_{v=1}^{a} \frac{v(v-1)}{2}} \raisepunct{.}
\label{equ: ww_rand}
\end{equation}

We also need to compare the effectiveness of these coefficients in assessing the randomness present in the investigations. To do this, we sample a number of tests from the 44 different ten times for each number of potential tests. This allows us to see how the different coefficients converge to the true coefficient value found by computing the metric over all tests.

To asses the quality of the results, we compare whether the performance is better under the hyperparameters or default reported by original implementations. The different parameter sets are given a rank by comparing the performance on every test. This is then averaged across every test, to give the Framework Comparison Rank (FCR) \cite{ugle2023leeney}. Demonstrating that failing to optimize hyperparameters properly can result in sub-optimal performance means that models that could have performed better with proper tuning may appear inferior. This affects decision-making and potentially leading to the adoption of sub-optimal solutions. In the real world, this can have costly and damaging consequences, and is especially critical in domains where model predictions impact decisions, such as healthcare, finance, and autonomous vehicles.

\subsection{Problem Definition}\label{section: problem_def}

The problem definition of community detection on attributed graphs is defined as follows. The graph, where $N$ is the number of nodes in the graph, is represented as $G = (A, X)$, with the relational information of nodes modelled by the adjacency matrix $A \in \mathbb{R}^{N \times N}$. Given a set of nodes $V$ and a set of edges $E$, let $e_{i, j} = (v_i, v_j) \in E$ denote the edge that points from $v_j$ to $v_i$. The graph is considered weighted so, the adjacency matrix $ 0 < A_{i, j} \leq 1$ if $e_{i,j} \in E$  and $A_{i, j} = 0$ if $e_{i,j} \notin E$. Also given is a set of node features $X \in \mathbb{R}^{N \times d} $, where $d$ represents the number of different node attributes (or feature dimensions). The objective is to partition the graph $G$ into $k$ clusters such that nodes in each partition, or cluster, generally have similar structure and feature values. The only information typically given to the algorithms at training time is the number of clusters $k$ to partition the graph into. Hard clustering is assumed, where each community detection algorithm must assign each node a single community to which it belongs, such that $P \in \mathbb{R}^{N}$ and we evaluate the clusters associated with each node using the labels given with each dataset such that $L \in \mathbb{R}^{N}$. Metrics that compare to the ground truth labels are allowed to be used for early-stopping and for hyperparameter optimisation. 

\subsection{Models}

We consider a representative suite of GNNs, selected based on factors such as code availability and re-implementation time.
In addition to explicit community detection algorithms, we also consider those that can learn an unsupervised representation of data as there is previous research that applies vector-based clustering algorithms to the representations \cite{mcconville2021n2d}.  The following are GNN architectures that learn representations of attributed graphs without a comparison to any labels during the training process. All these methods have hyperparameters which control trade-offs in optimisation.

Deep Attentional Embedded Graph Clustering (DAEGC) uses a k-means target to self-supervise the clustering module to iteratively refine the clustering of node embeddings \cite{wang2019attributed}. Deep Modularity Networks (DMoN) uses GCNs to maximises a modularity based clustering objective to optimise cluster assignments by a spectral relaxation of the problem \cite{tsitsulin2020graph}. Neighborhood Contrast Framework for Attributed Graph Clustering (CAGC) \cite{NCAGC} is designed for attributed graph clustering with contrastive self-expression loss that selects positive/negative pairs from the data and contrasts representations with its k-nearest neighbours. Deep Graph Infomax (DGI) maximises mutual information between patch representations of sub-graphs and the corresponding high-level summaries \cite{velickovic2019deep}. GRAph Contrastive rEpresentation learning (GRACE) generates a corrupted view of the graph by removing edges and learns node representations by maximising agreement across two views \cite{zhu2020deep}. Contrastive Multi-View Representation Learning on Graphs (MVGRL) argues that the best employment of contrastive methods for graphs is achieved by contrasting encodings' from first-order neighbours and a general graph diffusion \cite{hassani2020contrastive}. Bootstrapped Graph Latents (BGRL) \cite{thakoor2021bootstrapgraph} uses a self-supervised bootstrap procedure by maintaining two graph encoders; the online one learns to predict the representations of the target encoder, which in itself is updated by an exponential moving average of the online encoder. SelfGNN \cite{kefato2021selfgnn} also uses this principal but uses augmentations of the feature space to train the network. Towards Unsupervised Deep Graph Structure Learning (SUBLIME) \cite{liu2022towards} an encoder with the bootstrapping principle applied over the feature space as well as a contrastive scheme between the nearest neighbours. Variational Graph AutoEncoder Reconstruction (VGAER) \cite{qiu2022VGAER} reconstructs a modularity distribution using a cross entropy based decoder from the encoding of a VGAE \cite{kipf2016variational}. 

\subsection{Hyperparameter Optimisation Procedure}

There are sweet spots of architecture combinations that are best for each dataset \cite{bergstra2012random} and the effects of not selecting hyperparameters (HPs) have been well documented. Choosing too wide of a HP interval or including uninformative HPs in the search space can have an adverse effect on tuning outcomes in the given budget \cite{yang2020hyperparameter}. Thus, a HPO is performed under feasible constraints in order to validate the hypothesis that HPO affects the comparison of methods. It has been shown that grid search is not suited for searching for HPs on a new dataset and that Bayesian approaches perform better than random \cite{bergstra2012random}. There are a variety of Bayesian methods that can be used for hyperparameter selection. One such is the Tree Parzen-Estimator (TPE) \cite{bergstra2013making} that can retain the conditionality of variables \cite{yang2020hyperparameter} and has been shown to be a good estimator given limited resources \cite{yuan2021systematic}. The multi-objective version of the TPE \cite{ozaki2020multiobjective} is used to explore the multiple metrics of performance investigated. Given a limited budget, the TPE is optimal as it allows the efficient exploration of parameters.

%\begin{minipage}[t]{0.5\textwidth}
\begin{center}
\begin{table}[!htbp]
\centering
\newcolumntype{C}{>{\centering\arraybackslash}X}
\begin{tabularx}{\linewidth-60pt}{CC}
\toprule
Resource & Associated Allocation  \\
\midrule
Optimiser & Adam \\
Learning Rate & $\{0.05, 0.01, 0.005, 0.001, 0.0005, 0.0001 \}$ \\
Weight Decay & $\{0.05, 0.005, 0.0005, 0.0\}$ \\
Max Epochs & $5000$ \\
Patience & $\{25, 100, 500, 1000\}$ \\
Max Hyperparameter Trials & $300$ \\
Seeds & $\{42, 24, 976, 12345, 98765, 7, 856, 90, 672, 785\}$ \\
Training/Validation Split & $0.8$ \\ 
Train/Testing Split & $0.8$ \\
\bottomrule
\end{tabularx}
\caption{Resources are allocated an investigation, those detailed are shared across all investigations. Algorithms that are designed to benefit from a small number of HPs should perform better as they can search more of the space within the given budget. All models are trained with 1x 2080 Ti GPU on a server with 12GB of RAM, and a 16core Xeon CPU. \label{tab: resources} }
\end{table}
\end{center}

In this framework, a modification to the nested cross validation procedure is used to match reasonable computational budgets, which is to optimise hyperparameters on the first seed tested on and use those hyperparameters on the other seeds. Additionally, it it beneficial to establish a common resource allocation such as number of epochs allowed for in training or number of hyperparameter trials. Ensuring the same resources are used in all investigations means that the relatively underfunded researchers can participate in the field, democratising the access to contribution. Conversely, this also means that highly funded researchers cannot bias their results by exploiting the resources they have available. 

\begin{table}[!htbp] 
\newcolumntype{C}{>{\centering\arraybackslash}X}
\centering
\begin{tabularx}{\linewidth-60pt}{CCCCCCC}
\toprule
\textbf{\Longunderstack{\\Datasets}}  & \Longunderstack{\\Nodes} & \Longunderstack{\\Edges} & \Longunderstack{\\Features} & \Longunderstack{\\Classes} & \Longunderstack{Average\\Clustering\\ Coefficient} & \Longunderstack{Mean\\Closeness\\ Centrality} \\
\midrule
AMAC \cite{he2016ups} & 13752 & 160124 & 767 & 10 & 0.157 & 0.264 \\
AMAP \cite{he2016ups} & 7650 & 238163 & 745 & 8 & 0.404 & 0.242 \\
BAT \cite{deep_graph_clustering_survey} & 131 & 2077 & 81 & 4 & 0.636 & 0.469 \\
CiteSeer  \cite{giles1998citeseer} & 3327 & 9104 & 3703 & 6 & 0.141 & 0.045 \\
Cora \cite{mccallum2000automating} & 2708 & 10556 & 1433 & 7 & 0.241 & 0.137 \\
DBLP \cite{tang2008arnetminer} & 4057 & 7056 & 334 & 4 & 0.177 & 0.026 \\
EAT \cite{deep_graph_clustering_survey} & 399 & 11988 & 203 & 4 & 0.539 & 0.441 \\
UAT \cite{deep_graph_clustering_survey} & 1190 & 27198 & 239 & 4 & 0.501 & 0.332 \\
Texas \cite{craven1998learning} & 183 & 325 & 1703 & 5 & 0.198 & 0.344 \\
Wisc \cite{craven1998learning} & 251 & 515 & 1703 & 5 & 0.208 & 0.32 \\
Cornell \cite{craven1998learning} & 183 & 298 & 1703 & 5 & 0.167 & 0.326 \\
\bottomrule
\end{tabularx}
\caption{The datasets and associated statistics. \label{tab:stats}  }
\end{table}

\newpage
\subsection{Suite of Tests}
A test of an algorithm in this framework is the performance under a metric on a dataset. Algorithms are ranked on each test on every random seed used. For evaluation purposes, some metrics require the ground truth, and others do not, although regardless, this knowledge is not used during the training itself. Macro F1-score (F1) is calculated to ensure that evaluations are not biased by the label distribution, as communities sizes are often imbalanced. Normalised mutual information (NMI) is also used, which is the amount of information that can extract from one distribution with respect to a second. 

For unsupervised metrics, modularity and conductance are selected. Modularity quantifies the deviation of the clustering from what would be observed in expectation under a random graph. Conductance is the proportion of total edge volume that points outside the cluster. These two metrics are unsupervised as they are calculated using the predicted cluster label and the adjacency matrix, without using any ground truth. Many metrics are used in the framework as they align with specific objectives and ensures that evaluations reflect a clear and understandable assessment of performance. 

Generalisation of performance on one dataset can often not be statistically valid and lead to overfitting on a particular benchmark \cite{salzberg1997comparing}, hence multiple are used in this investigation. To fairly compare different GNN architectures, a range of graph topologies are used to fairly represent potential applications. Each dataset can be summarised by commonly used graph statistics: the average clustering coefficient \cite{watts1998collective} and closeness centrality \cite{wasserman1994social}. The former is the proportion of all the connections that exist in a nodes neighbourhood compared to a fully connected neighbourhood, averaged across all nodes. The latter is the reciprocal of the mean shortest path distance from all other nodes in the graph. All datasets are publicly available and have been used previously in GNN research \cite{deep_graph_clustering_survey}. 

Using many datasets for evaluation means that dataset bias is mitigated, which means that the framework is better at assessing the generalisation capability of models to different datasets. These datasets are detailed in Table~\ref{tab:stats}, the following is a brief summary: Cora \cite{mccallum2000automating}, CiteSeer \cite{giles1998citeseer}, DBLP \cite{tang2008arnetminer} are graphs of academic publications from various sources with the features coming from words in publications and connectivity from citations. AMAC and AMAP are extracted from the Amazon co-purchase graph \cite{he2016ups}. Texas, Wisc and Cornell are extracted from web pages from computer science departments of various universities \cite{craven1998learning}. UAT, EAT and BAT contain airport activity data collected from the National Civil Aviation Agency, Statistical Office of the European Union and Bureau of Transportation Statistics \cite{deep_graph_clustering_survey}.

\begin{figure}[!htbp] 
\centering
\includegraphics[width=15.5cm,height=0.85\textheight,keepaspectratio]{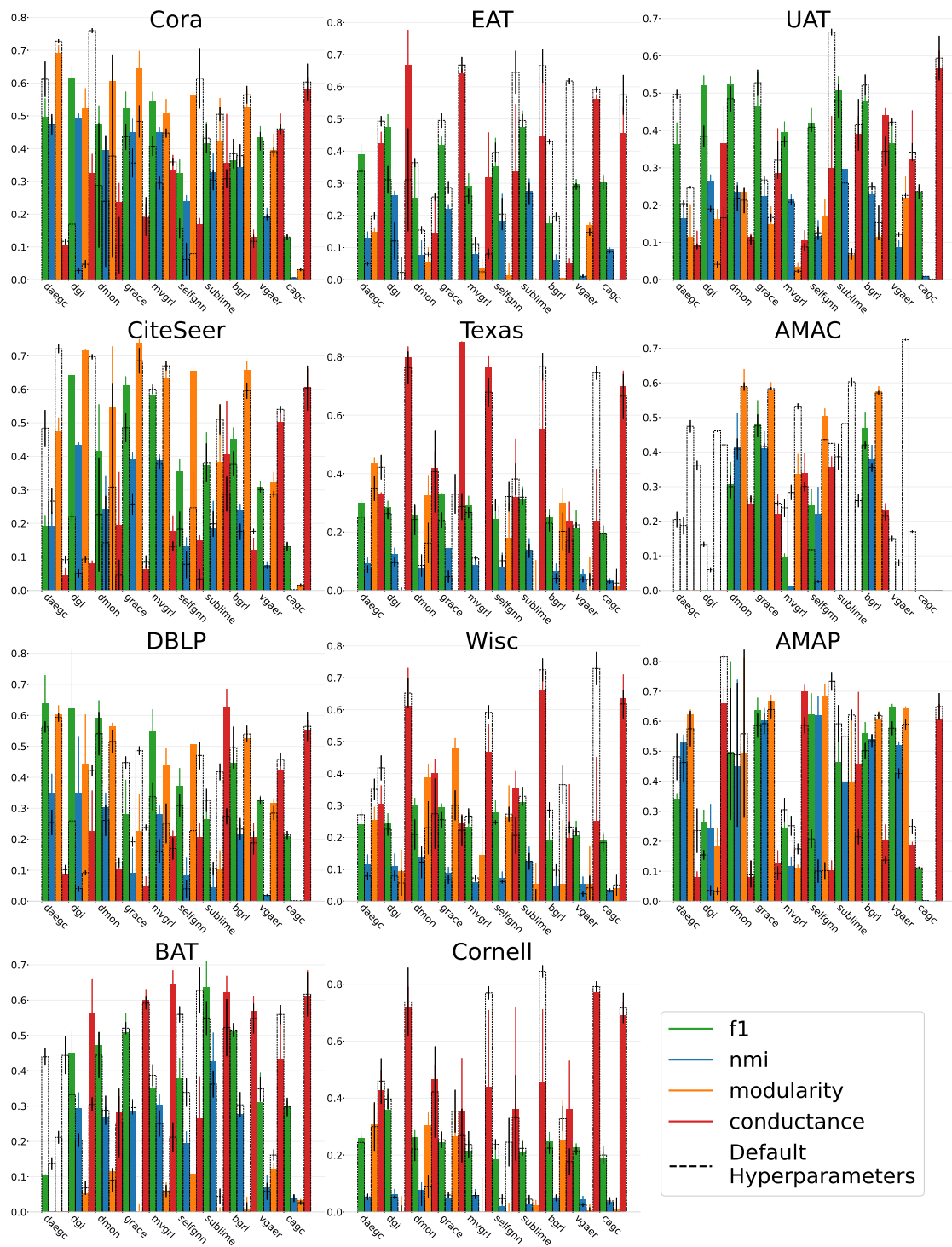}
\caption{The average performance and standard deviation of each metric averaged over every seed tested on for all methods on all datasets. The hyperparameter investigation under our framework is shown in colour compared with the default hyperparameters in dashed boxes. The lower the value for Conductance is better. Out of memory occurrences on happened on the AMAC dataset with the following algorithms during HPO: DAEGC, SUBLIME, DGI, CAGC, VGAER and CAGC under the default HPs.\label{fig: results_fig}} 
\end{figure}
\newpage

%%%%%%%%%%%%%%%%%%%%%%%%%%%%%%%%%%%%%%%%%%
\begin{table}[!htbp] 
\centering
\newcolumntype{C}{>{\centering\arraybackslash}X}
\begin{tabularx}{\linewidth-60pt}{CCC}
\toprule
Results & Default & HPO \\
\midrule
\Longunderstack{Framework \\ Comparison Rank} & $1.829$ & $1.171$ \\
\Longunderstack{$W$ Randomness\\ Coefficient} & $0.476$ & $0.489$ \\
N Ties & 10 & 200 \\
\Longunderstack{$W$ Randomness\\w/ Mean Ties} & $0.150$ & $0.245$\\
\Longunderstack{Tied $W_t$ Randomness} & $0.150$ & $0.229$ \\
\Longunderstack{$W_w$ Wasserstein Randomness} & $0.072$ & $0.127$ \\
\bottomrule
\end{tabularx}
\caption{Here we show the quantification of intra framework consistency using the $W$ Randomness Coefficient, $W$ Randomness with Mean Ties, Tied $W$ Randomness, $W$ Wasserstein Randomness and inter framework disparity using the Framework Comparison Rank. Low values for the Framework Comparison Rank and all $W$ Randomness Coefficient's are preferred. \label{tab: metric_stats}}
\end{table}

\begin{figure}[!htbp] 
\centering
\includegraphics[width=13.5cm,height=11cm,keepaspectratio]{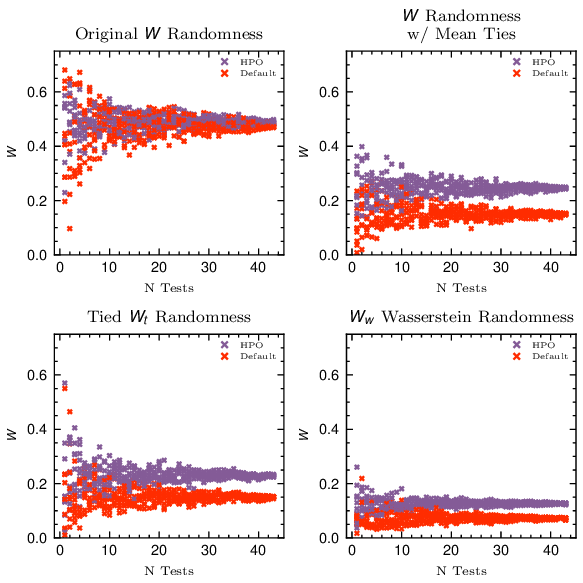}
\caption{Here the different metrics of quantifying randomness are compared against samples of the original testing space. The randomness in rankings over samples of experiments using the default hyperparameters is compared against the HPO results. The distinction between the original $W$ Randomness Coefficient proposed by \citet{ugle2023leeney} and the other metrics of randomness is that ties are settled by taking the mean of the ranks rather than assigning the lowest rank to all algorithms. \label{fig: distribution_fig}} 
\end{figure}

\section{Evaluation and Discussion}

The Framework Comparison Rank is the average rank when comparing performance of the parameters found through hyperparameter optimisation versus the default values. From Table~\ref{tab: metric_stats} it can be seen that Framework Comparison Rank indicates that the hyperparameters that are optimised on average perform better. This validates the hypothesis that the hyperparameter optimisation significantly impacts the evaluation of GNN based approaches to community detection. 

The results of the hyperparameter optimisation and the default parameters are visualised in Figure~\ref{fig: results_fig}. From this, we can see the difference in performance under both sets of hyperparameters. On some datasets the default parameters work better than those optimised, which means that under the reasonable budget assumed in this framework they are not reproducible. This adds strength to the claim that using the default parameters is not credible. Without validation that these parameters can be recovered, results cannot be trusted and mistakes are harder to identify. 
On the other side of this, often the hyperparameter optimisation performs better than the default values. Algorithms were published knowing these parameters can be tuned to better performance. Using a reasonable resources, the performance can be increased, which means that without the optimisation procedure, inaccurate or misleading comparisons are propagated. Reproducible findings are the solid foundation that allows us to build on previous work and is a necessity for scientific validity. 

The $W$ Randomness Coefficient quantifies the consistency of rankings over the different random seeds tested on, averaged over the suite of tests in the framework. With less deviation of prediction under the presence of randomness, an evaluation finds a more confident assessment of the best algorithm. A marginally higher $W$ value using the optimised hyperparameters indicates that the default parameters are more consistent over randomness. There is little difference in the coefficients for the original $W$ randomness coefficient, which is potentially due to the fact that the default parameters have been evaluated with a consistent approach to model selection and constant resource allocation to training time. However, when we account for the number of tied ranks, we find that the HPO has more tied ranks over the investigation, due to out of memory or optimial conductance values. This may be because the hyperparamter optimisation leads to better results, reducing the difference between algorithms. This is supported by the $W$ randomness coefficients that account for ties, showing that the HPO is less consistent across randomness. By assessing the $W$ Randomness coefficient we can reduce the impact of biased evaluation procedures. However it is important to explain the reasons behind why randomness might be affecting the results, the higher $W$ for the hyperparameter analysis might be because algorithms increase performance under the HPO. Therefore, when all methods are fairly evaluated under a HPO the the spread of rankings may overlap more, which isn't explictly negative. With this coefficient, researchers can quantify how reliable their results are, and therefore the usability in real-world applications. This sets the baseline for consistency in evaluation procedure and allows better understanding of relative method performance.

In Figure~\ref{fig: distribution_fig} we see the convergence of each metric as the number of tests increases. There is significant overlap between the default and HPO investigation. When we use the original $W$ randomness coefficient but change how tied ranks are dealt with, by taking the mean of the ties, we can see that the two distributions overlap less. This demonstrates that how ties are dealt with does change quantification of randomness. Comparing this to the $W_t$ randomness coefficient, we can see that this metric does not affect the measurement of randomness at the scale that is carried out herein. However, when comparing to the $W_w$, it is clear that this coefficient has significantly less spread for low number of tests and converges quicker to the true value as measured by the full breadth of the investigation. Therefore, this metric can be used to provide a more accurate assessment of the amount of randomness in the investigation and the robustness of the methods evaluated.

Given different computational resources performance rankings will vary. Future iterations of the framework should experiment with the the number of trials and impact of over-reliance on specific seeds or extending the hyperparameter options to a continuous distribution. Additionally, finding the best general algorithm will have to include a wide range of different topologies or sizes of graphs that are not looked at, neither do we explore other feature space landscapes or class distributions.

%%%%%%%%%%%%%%%%%%%%%%%%%%%%%%%%%%%%%%%%%%
\section{Conclusion}

In this work we demonstrate flaws with how GNN based community detection methods are currently evaluated, leading to potentially misleading and confusing conclusions. To address this, a framework compared for providing a more consistent and fair evaluation for GNN community detection. We provide further insight that consistent HPO is key at this task by quantifying the difference in performance from HPO to reported values. Finally, different metrics of assessing the consistency of rankings under the presence of randomness are compared. It is found that the $W$ Wasserstein randomness coefficient is the most robust to samples of the full test investigation.

\newpage
\bibliographystyle{plainnat}
\bibliography{ref}

\end{document}